\documentclass[10pt,twocolumn,letterpaper]{article}

\usepackage{cvpr}
\usepackage{times}
\usepackage{epsfig}
\usepackage{graphicx}
\usepackage{amsmath}
\usepackage{amssymb}
\usepackage{booktabs}
\usepackage{multirow}
\usepackage[table,xcdraw]{xcolor}

\usepackage[breaklinks=true,bookmarks=false,colorlinks]{hyperref}

\cvprfinalcopy 

\def\cvprPaperID{9469} 

\begin{document}

\title{Object Relational Graph with Teacher-Recommended Learning \\for Video Captioning}

\author{\textbf{Ziqi Zhang}\textsuperscript{1,3}\thanks{Equal contribution.}, 
		\textbf{Yaya Shi}\textsuperscript{2}\footnotemark[1],
		\textbf{Chunfeng Yuan}\textsuperscript{1}\thanks{Corresponding author.},
		\textbf{Bing Li}\textsuperscript{1,6,7},
		\textbf{Peijin Wang}\textsuperscript{3,5},
		\textbf{Weiming Hu}\textsuperscript{1,3,4},
		\textbf{Zhengjun Zha}\textsuperscript{2}\\
 \textsuperscript{1}National Laboratory of Pattern Recognition, CASIA\\\textsuperscript{2}University of Science and Technology of China \textsuperscript{3}University of Chinese Academy of Sciences\\
 \textsuperscript{4}Center for Excellence in Brain Science and Intelligence Technology, CAS\\
 \textsuperscript{5}Aerospace Information Research Institute, CAS \textsuperscript{6}PeopleAI, Inc.\\
  \textsuperscript{7}State Key Laboratory of Communication Content Cognition, People's Daily Online\\
 {\tt\small
  \{zhangziqi2017\}@ia.ac.cn,\{{shiyaya,zhazj}\}@mail.ustc.edu.cn,\{cfyuan,bli,wmhu\}@nlpr.ia.ac.cn}
 }

\maketitle

\begin{abstract} 
	Taking full advantage of the information from both vision and language is critical for the video captioning task. Existing models lack adequate visual representation due to the neglect of interaction between object, and sufficient training for content-related words due to long-tailed problems. In this paper, we propose a complete video captioning system including both a novel model and an effective training strategy. Specifically, we propose an object relational graph (ORG) based encoder, which captures more detailed interaction features to enrich visual representation. Meanwhile, we design a teacher-recommended learning (TRL) method to make full use of the successful external language model (ELM) to integrate the abundant linguistic knowledge into the caption model. The ELM generates more semantically similar word proposals which extend the ground-truth words used for training to deal with the long-tailed problem. Experimental evaluations on three benchmarks: MSVD, MSR-VTT and VATEX show the proposed ORG-TRL system achieves state-of-the-art performance. Extensive ablation studies and visualizations illustrate the effectiveness of our system.  
\end{abstract}

\section{Introduction}
Video captioning aims to generate natural language descriptions automatically according to the visual information of given videos. There are many wonderful visions of video captioning such as blind assistance and autopilot assistance. Video captioning needs to consider both spatial appearance and temporal dynamics of video contents, which is a promising and challenging task. The key problems in this task are twofold: how to extract discriminative features to represent the contents of videos, and how to leverage the existing visual features to match the  corresponding captioning corpus. The ultimate aim is to cross the gap between vision and language. 

For vision representation, previous works~\cite{Yao2015, Yu2016, Pan2016, Pan2016b, Wang2018a} always leverage appearance features of keyframes and motion features of segments to represent video contents. These features extract global information and hard to capture the detailed temporal dynamics of objects in the video. The most recent works~\cite{Zhang2019,Hu2019} apply a pretrained object detector to obtain some object proposals in each keyframe and use spatial/temporal attention mechanisms to fuse object features. However, they neglect the relationship between objects in the temporal and spatial domains. Some researches in the field of Visual Question Answering, Image Captioning even Action Recognition demonstrate that the relationship between objects is vital, which also plays an important role in generating a more detailed and diverse description for a video.

\begin{figure}
	\centering
	\includegraphics[width=1\linewidth]{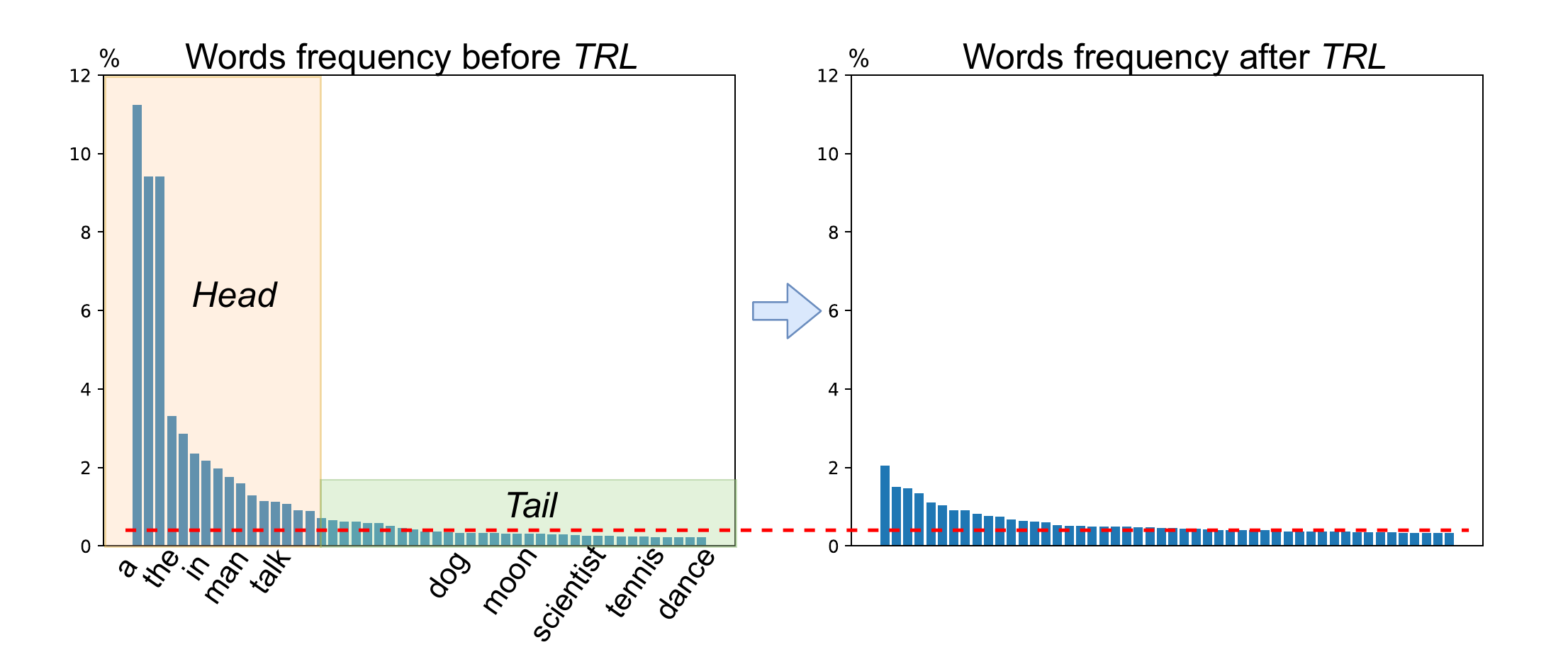}
	\caption{Long-tailed problem of the caption corpus. The top-50 words frequency of MSR-VTT are shown. Under the guidance of TRL, more potential content-specific words are exposed to the caption model. Compared with the words frequency before, the words in tail region get an overall boosting.}
	\label{fig:long-tail}
	\vspace{-0.6cm}
\end{figure}

For sentence generation, according to statistics of word frequency in caption corpus, it is found that the majority of words are function words and common words \eg ``the'' and ``man'', which are far more than the real content-specific words in number. This is the so-called ``long-tailed'' problem, as shown in Fig.\ref{fig:long-tail}. This problem will cause insufficient training for a large number of meaningful words. Although the long-tailed problem can be relieved by giving different weights to different words~\cite{Dong2018}, it can not be solved fundamentally. Furthermore, a caption model should not only comprehend visual information but also grasp linguistic ability using such a small number of samples, which is a so heavy task! Why not employ a ready-made ELM \eg BERT~\cite{Devlin2019} or GPT~\cite{radford2018improving} as a teacher, to directly impart linguistic knowledge to the caption model, to mitigate the problem caused by insufficient samples.

In this paper, we propose a novel model with the assistance of an original training strategy to deal with above two issues for video captioning:
1) We construct a learnable ORG to fully explore the spatial and temporal relationships between objects. With the help of graph convolutional networks(GCNs)~\cite{Kipf2017}, object representations can be enhanced during the process of relational reasoning. Specifically, we explore two kinds of graphs: the partial object relational graph (P-ORG) connects objects in the same frame, and the complete object relational graph (C-ORG) builds a connection for all the objects in video. Scaled dot-product is utilized to implicitly compute relationships between each object, which is learnable during training. Finally, object features are updated by GCNs to be more informative features.
2) Generally, the caption model is forced to learn the ground-truth word at each training step, so we call this process as the teacher-enforced learning (TEL) and these words as “hard target”. However, TEL doesn't consider the long-tailed problem. Therefore, we propose a TRL method, which makes full use of external language model (ELM) to generate some word proposals according to the prediction probability of current ground-truth words. These proposals are called “soft targets”, which are often semantically similar with the ground-truth words and extended them. Specifically, the ELM is off-line well-trained on a large-scale external corpus, and it is employed as an experienced “teacher”, who has contained a wealth of linguistic knowledge. By contrast, the caption model can be regarded as a “student”. Under the guidance of TRL, excellent linguistic knowledge from ELM is transformed into the caption model.

The contributions of this work can be summarized as following: 1) We construct novel ORGs to connect each object in video and utilizes GCNs to achieve relational reasoning, which enrich the representation of detailed objects further. 2) The TRL is proposed as a supplement of the TEL, to integrate linguistic knowledge from an ELM to the caption model. Several times words are trained at each time step more than before. It's effective to relieve long-tailed problem and improve generalization of the caption model. 3) Our model achieves state-of-the-art performances on three benchmarks: MSVD, MSR-VTT, and newly VATEX.

\section{Related Works}
\textbf{Video Captioning.} Recent researches mainly focus on sequence-learning based methods~\cite{Venugopalan2015, Yao2015, Yu2016, Pan2016, Pan2016b, Wang2018a, Pei2019}, which adopt encoder-decoder structure. Yao \etal~\cite{Yao2015} propose a temporal attention mechanism to dynamically summarize the visual features. Wang \etal~\cite{Wang2018a} try to enhance the quality of generated captions by reproducing the frame features from decoding hidden states. More recently, there are some researches concerning the object-level information~\cite{Yang2017, Zhang2019, Hu2019}. Zhang \etal~\cite{Zhang2019} use a bidirectional temporal graph to capture detailed temporal dynamics for the salient objects in the video. Hu \etal~\cite{Hu2019} use two-layers stacked LSTM as an encoder to construct the temporal structure at frame-level and object-level successively.

However, these methods mainly work on the global information or temporal structure of salient objects without considering the interactions between each object in frames. In this work, we propose a graph-based approach, which constructs a temporal-spatial graph on all the objects in a video to enhance object-level representation.

\textbf{ Visual Relational Reasoning.} Some researches have shown that visual relational reasoning is effective for computer vision tasks, such as Image Captioning~\cite{Yao2018, Yang2019}, VQA~\cite{Norcliffe-Brown2018, Narasimhan2018, Li2019} and Action Recognition~\cite{Wang2018, Wang2018b}. Yao \etal~\cite{Yao2018} exploit predefined semantic relations learned from the scene graph parsing task ~\cite{Zellers2018} and embed the graph structure into vector representations by using a modified GCN. Li \etal~\cite{Li2019} use both explicit graph and learnable implicit graph to enrich image representation and apply GAT ~\cite{Velickovic2017} to update relations in attentive weight. Wang \etal~\cite{Wang2018b} compute both implicit similarity relation and relative positional relation of each object in the video, and then apply GCNs to perform reasoning. There are few efforts utilizing relational reasoning for video captioning. 

\begin{figure*}
	\centering
	\includegraphics[width=1\linewidth]{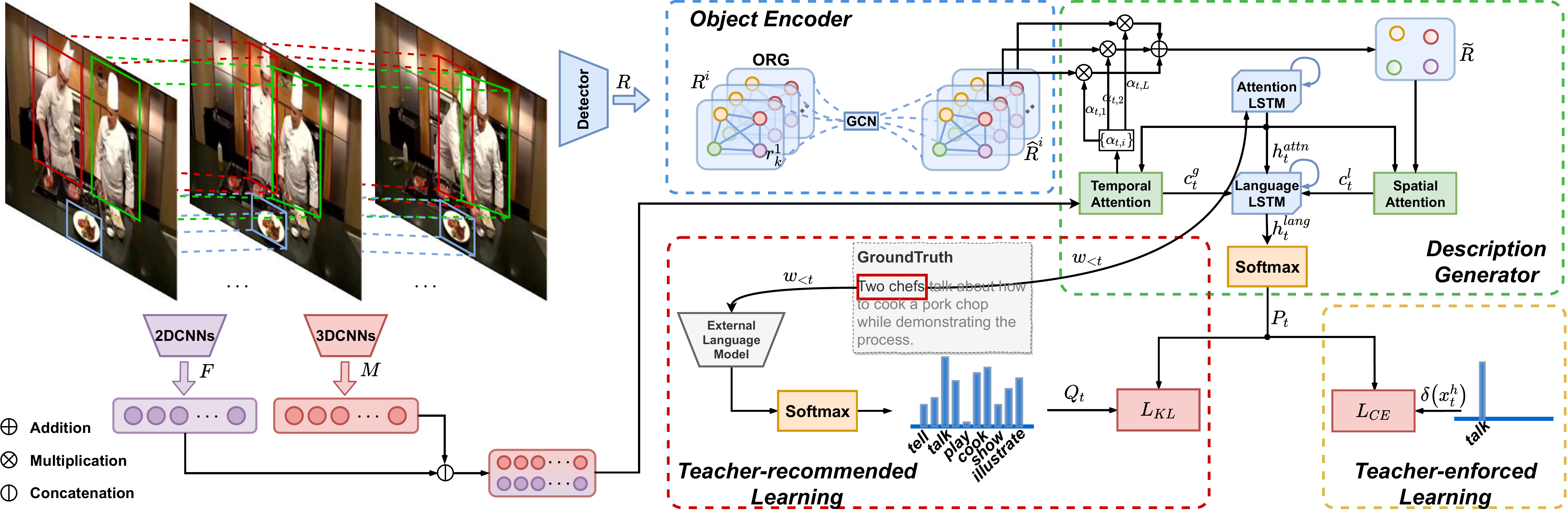}
	\caption{The overview of our proposed ORG-TRL system. It mainly consists of the ORG based object encoder presented in the top-left box, and the hierarchical decoder with temporal/spatial attention in the top-right box. Our model is under the co-guidance of the novel TRL in the bottom-left box and the common TEL in the bottom-right. It also illustrates a virtual example during training: when $t=3$, the TEL forces the model to learn ``talk'', but the TRL recommends the model to learn more words via the knowledge from ELM.}
	\label{fig:framework}
	\vspace{-0.5cm}
\end{figure*}

\textbf{External Language Model for Seq2Seq Generation Tasks.} ELM has been applied to many natural language generation tasks such as neural machine translation (NMT) and automatic speech recognition (ASR). An early attempt to use ELM for NMT in~\cite{Guelcehre2015} is also known as \textit{shallow fusion} and \textit{deep fusion}. Kannan \etal~\cite{Kannan2018} fully explore the behavior of shallow fusion with different ELMs and test them on a large-scale ASR task. Sriram \etal~\cite{Sriram2018} propose \textit{cold fusion} to improve ASR performance.

These above fusion methods illustrate promising performance but also have some limitations. Shallow fusion may bring bias when output logits are used directly because of the difference in the data distribution between language model and task model. Deep fusion also needs ELM during inference and cold fusion relies on additional gating mechanisms and networks, which will bring heavy calculations and complexity to the task model. In comparison, our introduced TRL method only calculates the KL divergence between ``soft targets'' and output distribution of task model during training which can well overcome above-mentioned limitations.

\section{Methodology}

Fig.\ref{fig:framework} illustrates the overview of our system. An encoder-decoder framework is followed. Appearance, motion and detailed objects features are extracted by diverse networks. Specifically, we construct a graph based object encoder whose core is a learnable object relational graph (ORG), which can learn the interaction among different objects dynamically. The description generator generates each word by steps, with attentively aggregating visual features in space and time. For the learning process, not only normal teacher-enforced learning (TEL) but also proposed teacher-recommended learning (TRL) strategy are leveraged to learn task-specific knowledge and external linguistic knowledge separately.

\subsection{Object Relational Graph based Visual Encoder}

Formally, given a sequence of video frames, we uniformly extract $ T $ frames as keyframes, and collect a short-range video frames around keyframes as segments which reflects the temporal dynamics of a video.
The pretrained 2D CNNs and 3D CNNs are employed to extract the appearance features $ \mathcal{F}=\left\lbrace f_i \right\rbrace $ of each keyframe and motion features $ \mathcal{M} = \left\lbrace m_i \right\rbrace $ of each segment separately, where $f_i $ and $ m_i $ denote the features of the $ i_{th} $ frame and segment respectively; $i = 1,\dots,L $; $ L $ denotes the number of keyframes.

People always describe an object based on its relationships with others in the video. In order to get the detailed object representations, the pretrained object detector is applied to capture several class-agnostic object proposals in each keyframe and extract their features $ \mathcal{R}^{i} = \left\lbrace r_k^i \right\rbrace, i = 1,\dots,L, k = 1,\dots,N $, where $ r_k^i $ represents the $ k_{th} $ object feature in $ i_{th}$ keyframe, $L$ is the number of keyframes and $ N $ is the number of objects in each frame. These original object features are independent, and they have no interaction with each other in time and space.

To learn the relation message from surrounding objects, we define a relational graph for a object set and then use it to update the object features. Specifically, given $K$ objects, each object is considered as a node. Let $R \in \mathbb{R}^{K\times d}$ denote $K$ object nodes with $d$ dimensional feature, and $A \in \mathbb{R}^{K \times K}$ denote the relation coefficient matrix between $K$ nodes. We define $A$ as: 
\begin{equation}\label{key}
A = \phi(R)\cdot\psi(R)^T
\end{equation}
\begin{equation}\label{key}
\phi(R) = R\cdot W_i+b_i, \psi(R) = R\cdot W_j+b_j
\end{equation}
where $W_i,W_j \in \mathbb{R}^{d\times d^{'}}$ and $b_i \in \mathbb{R}^d,b_j \in \mathbb{R}^{d^{'}}$ are learnable parameters. Subsequently, $A$ is normalized to make the sum of edges, connecting to the same node, equals to 1:
\begin{equation}\label{key}
\hat{A} = softmax(A, dim=1)
\end{equation}
where $\hat{A}$ can be seen as how much information the center object gets from the surrounding objects. We apply GCNs to perform relational reasoning, then original objects features $R$ are updated to $\hat{R}$:
\begin{equation}\label{key}
\hat{R} = \hat{A}\cdot R\cdot W_r
\end{equation}
where $\hat{R} \in \mathbb{R}^{K\times d}$ is enhanced object features with interaction message between objects, and $W_r \in \mathbb{R}^{d\times d}$ is learnable parameters.

\begin{figure}
	\begin{center}
	\includegraphics[width=1\linewidth]{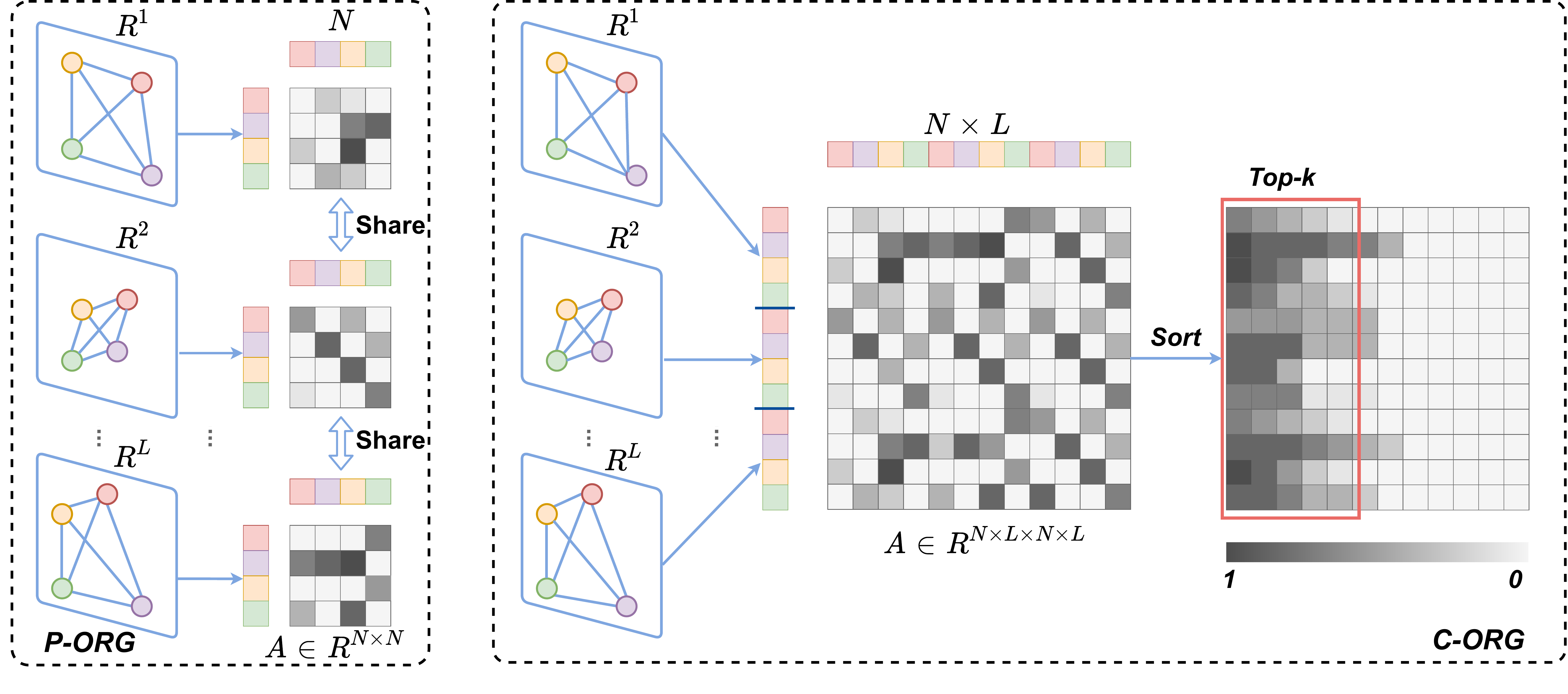}
	\end{center}
	\caption{The diagrams of the proposed P-ORG and C-ORG. Each colored square represents the vector of the object. $A$ is the relational coefficient matrix. }
	\label{fig:org}
	\vspace{-0.5cm}
\end{figure}

We explore two kinds of relational graphs as shown in Fig.\ref{fig:org}, the P-ORG and the C-ORG. Specifically, the P-ORG only build the relationship between $N$ objects in the same frame thus a $A \in \mathbb{R}^{N \times N}$ relational graph is constructed. Note that learnable parameters of relational graph are shared with all $L$ frames. Although object proposals appearing in different frames may belong to the same entity, they are considered as different nodes because of diverse states. Meanwhile, the C-ORG constructs a complete graph $A \in \mathbb{R}^{(N\times L)\times (N\times L)}$ which connects each object with all the other $N \times L$ objects in the video. It's noisy to directly connect center node with all $N \times L$ nodes, thus we select top-k corresponding nodes to connect.

Finally, the enhanced object features are computed by performing relational reasoning. They are together with appearance and motion features to sufficiently present videos.

\begin{table*}[]
	\begin{center}
	\resizebox{\textwidth}{!}{%
		\begin{tabular}{c|ccccccccccccc}
			\hline
			t=4 & A & woman & with & \cellcolor[HTML]{FFE6CC}green &  &  &  &  &  &  &  &  &  \\ \hline
			\cellcolor[HTML]{FFE6CC}soft targets & long & blonde & short & curly & brown & black & red & dark & white & pink &  &  &  \\
			\cellcolor[HTML]{FFE6CC}probability & 0.047 & 0.025 & 0.020 & 0.019 & 0.017 & 0.016 & 0.015 & 0.014 & 0.011 & 0.011 & \multirow{-2}{*}{...} &  &  \\ \hline
			t=6 & A & woman & with & green & hair & \cellcolor[HTML]{F8CECC}teaching &  &  &  &  &  &  &  \\ \hline
			\cellcolor[HTML]{F8CECC}soft targets & shows & showing & demonstrates & demonstrating & explains & explaining & teaches & teaching & tells & describes &  &  &  \\
			\cellcolor[HTML]{F8CECC}probability & 0.086 & 0.066 & 0.065 & 0.033 & 0.023 & 0.015 & 0.014 & 0.014 & 0.013 & 0.012 & \multirow{-2}{*}{...} &  &  \\ \hline
			t=13 & A & woman & with & green & hair & teaching & how & to & trim & flowers & in & a & \cellcolor[HTML]{E1D5E7}vase \\ \hline
			\cellcolor[HTML]{E1D5E7}soft targets & vase & garden & greenhouse & yard & room & field & tree & backyard & flower & bouquet &  &  &  \\
			\cellcolor[HTML]{E1D5E7}probability & 0.074 & 0.051 & 0.015 & 0.015 & 0.014 & 0.010 & 0.0095 & 0.0081 & 0.0068 & 0.0061 & \multirow{-2}{*}{...} &  &  \\ \hline
		\end{tabular}%
	}
	\end{center}
	\caption{An example of ``soft targets'' and ``hard target'' (colored words) at three positions of the same sentence. Given the words before ``hard targets'', the ELM generate 10 ``soft targets'' and their probabilities in descending order.}
	\label{tab:elm}
	\vspace{-0.5cm}
\end{table*}

\subsection{Description Generation}

After getting the sufficient video features, we propose a hierarchical decoder with a temporal-spatial attention module to generate linguistic descriptions by steps. The hierarchical decoder consists of the Attention LSTM and the Language LSTM. 

Firstly, the Attention LSTM is to summarize current semantics $h_t^{attn}$ according to the history hidden state $h_{t-1}^{lang}$ of Language LSTM, concatenated with mean-pooled global video feature $\bar{v}=\frac{1}{L}\sum v_i$ and the previous word $w_{t-1}$ at the decoding step $t$:
\begin{equation}\label{key}
h_t^{attn} = \textbf{LSTM}^{attn}\left(\left[ \bar{v}, W_ew_{t-1},  h_{t-1}^{lang}\right] ; h_{t-1}^{attn} \right)
\end{equation}
where $v_i = [f_i,m_i],f_i \in \mathcal{F},m_i \in \mathcal{M} $, is the concatenation of appearance feature and motion feature, $W_e$ is the learnable word embedding matrix. 

Following the current semantics $h_t^{attn}$, the temporal attention module dynamically decides when (frames) to attend, and abstracts the global context features $c_t^g$:
\[ c_t^g = \sum_{i=1}^{L}{\alpha_{t,i}v_i} \]
\begin{equation}\label{key}
\alpha_{t,i} = softmax\left( w_a^T tanh\left( W_av_i + U_ah_t^{attn}\right)\right)
\end{equation}
where $\alpha_{t,i}$ is the weight of the $i_{th}$ global feature at the $t_{th}$ decoding step; $L$ is the number of keyframes; $w_a,W_a$ and $U_a$ are learnable parameters.

For local object feature, objects in different frames are firstly aligned to merge together, and then the spatial attention module chooses which objects should be focused on. We use a simple but effective method to align objects in different frames. The process is shown on the left pictures in Fig.\ref{fig:framework}, and the dotted line trajectories present the objects alignment. We set objects in the first frame as anchors, and define $sim_i(j,j^{'})$ as the cosine distance between the $j_{th}$ object in anchor frame and the $j_{th}^{'}$ in $i_{th}$ frame:
\begin{equation}\label{key}
sim_i\left(j,j^{'}\right)  = cos\left( r_j^{1}, r_{j^{'}}^{i}\right)
\end{equation}
where $j,j^{'}=1,\dots,N;i=2,\dots,L$. Considering the similarity between two objects themselves, we use original object features $\mathcal{R}$ to calculate similarity rather than enhanced features $\hat{\mathcal{R}}$. The object in each frame is aligned to the anchors according to the maximum similarity. These aligned objects ideally belong to the same entity. Enhanced features $\hat{\mathcal{R}}$, following the group of aligned objects, are weighted sum by $\{\alpha_{t,i}\},i=1,\dots,L$. In this way, objects in different frames are merged into one frame as local aligned features $\tilde{R}$ according to alignment operation and temporal attention.

Then, the spatial attention module decides where (objects) to attend, and abstracts local context feature $c_t^l$:
\[ c_t^l = \sum_{j=1}^{N}{\beta_{t,j}u_j} \]
\begin{equation}\label{key}
\beta_{t,j} = softmax\left( w_b^T tanh\left( W_bu_j + U_bh_t^{attn}\right) \right)
\end{equation}
where $u_j \in \tilde{R}$ denotes one of the $N$ local aligned features; $w_b,W_b$ and $U_b$ are learnable parameters.

Finally, the Language LSTM summarizes both global and local context features to generate current hidden state $h_t^{lang}$. The probability distribution of the caption model $P_{t}$ is acquired, followed with a single layer perceptron and the softmax operation at decoding step $t$:
\begin{equation}\label{key}
h_t^{lang} = \textbf{LSTM}^{lang}\left( \left[ c_t^g,c_t^l,h_t^{attn}\right]; h_{t-1}^{lang} \right)
\end{equation}
\begin{equation}\label{key}
P_{t} = softmax(W_zh_t^{lang}+b_z)
\end{equation}
where $[\cdot,\cdot]$ denotes concatenation; $P_{t}$ is a $D$-dimensional vector of vocabulary size; $W_z$ and $b_z$ are learnable parameters.

\subsection{Teacher-recommended Learning via External Language Model}

For a sufficient training of content-specific words, the proposed model is jointly trained under the guidance of common TEL and proposed TRL.

As for conventional TEL process, the caption model is forced to generate the ground-truth word at each time step. This word is the so-called ``hard target'', which are expressed as $ \mathcal{X}_{hard} = \left\lbrace x_1^h,x_2^h,\dots,x_{T_s}^h \right\rbrace $, where $ x_t^h $ is one ground-truth word at the $ t_{th} $ decoding step; $T_s$ denotes training step in total of the given sentence. We refer to our designed caption model as CAP, and the output probability distribution of CAP is $P_t = CAP(w_{<t}|\theta_{CAP})$, where $w_{<t}$ is history words; $\theta_{CAP}$ stands all the parameters of the CAP. The training criterion is based on Cross-Entropy loss, only the probability corresponding to ground-truth participate in calculation:
\begin{equation}\label{key}
\mathcal{L}_{CE}(\theta) = - \sum_{t=1}^{T} \delta(x_t^h)^T \cdot logP_t
\end{equation}
where $\delta(d)\in \mathbb{R}^{D}$ denotes one-hot vector, and the value equals to 1 only at the word $d$ position; $P_t\in\mathbb{R}^{D}$ is the output distribution of the CAP; $x^h_t$ is the ``hard targets''. 

The TEL is lack of sufficient training for content-related words due to long-tailed problems. Therefore, we propose the TRL to integrate the knowledge from ELM. There are many ready-made models that can be employed as ELM \eg Bert and GPT. Suppose we got an ELM that has been well trained on a large scale monolingual corpus. When given the previous $t-1$ words $w_{<t}$, the probability distribution of ELM at time step $ t $ is:
\begin{equation}\label{key}
Q_t = ELM\left( w_{<t},T_e|\theta_{ELM}\right)
\end{equation}
where $Q_{t}\in \mathbb{R}^{D}$ is a $D$-dimensional vector representing the output distribution of ELM; $\theta_{ELM}$ are parameters of ELM which are fixed during the training phase of the CAP; $T_e$ is the temperature used to smooth output distribution. 

Generally, in order to transfer the knowledge from ELM to the CAP, it's easy to minimize the KL divergence between probability distribution of the CAP and the ELM during decoding step. To make $P_t$ fit $Q_t$, the KL divergence is formulated as:
\begin{equation}\label{key}
D_{KL}(Q_t||P_t) = -\sum_{d\in D}{Q_{t}^d \cdot log\frac{P_{t}^d}{Q_{t}^d}}
\end{equation}
where $P_{t}^d$ and $Q_{t}^d$ are the output probability of word $d$ in the CAP and the ELM respectively.

$Q_t$ is the probability distribution of all the words for task vocabulary, but most of the values($<10^{-4}$) are extremely small. These semantic irrelevant words may confuse the model and increase computation. Therefore we only extract top-k words as ``soft targets'':
\begin{equation}\label{key}
\mathcal{X}_{soft} = \left\lbrace \pmb{x}_1^s,\pmb{x}_2^s,\dots,\pmb{x}^s_{T_{s}} \right\rbrace
\end{equation}
where $\pmb{x}_t^s = \left\lbrace x_i^s|i=1,2,\dots,k\right\rbrace$ are a set of words in descending order of probability distribution $Q_{t}$ at the $t_{th}$ decoding step. Furthermore, the ELM is fixed while the CAP is training, so the KL-loss function is simplified as:
\begin{equation}\label{key}
\mathcal{L}_{KL}(\theta) = - \sum_{t=1}^{T}\sum_{d \in \pmb{x}_t^s} {Q_{t}^d \cdot log{P_{t}^d}}
\end{equation}

In most cases, ``hard target'' is concluded in ``soft targets'', because ELM is trained on the large-scale corpora. Tab.\ref{tab:elm} shows an example, our ELM can generate some syntactically correct and semantically reasonable proposals, which can be regarded as supplements to ground-truth word.

For the overall training process, our CAP is under the co-guidance of both TEL and TRL to learn task-specific knowledge and external linguistic knowledge separately. We set a trade-off parameter $\lambda \in [0,1]$ to balance the degree of TEL and TRL, thus the criterion of the whole system is shown as:
\begin{equation}\label{key}
\mathcal{L}(\theta) = \lambda \mathcal{L}_{KL}(\theta) + (1-\lambda)\mathcal{L}_{CE}(\theta)
\end{equation}

The TRL exposes a large number of potential words to the CAP. To some extent, it effectively alleviates the long-tailed problem of the caption training corpus. Moreover, there is no extra computational burden on sentence generation at inference time, because the TRL only participates in the training process of the CAP.

\section{Experiments}
In this section, we evaluate our proposed model on three datasets: MSVD~\cite{Chen2011}, MSR-VTT~\cite{Xu2016} and VATEX~\cite{Wang2019}, via four popular used metrics including BLEU-4~\cite{Papineni2002}, METEOR~\cite{Denkowski2014}, CIDEr~\cite{Vedantam2015} and ROUGE-L~\cite{Lin2004}. Our results are compared with state-of-the-art results, which demonstrate the effectiveness of our methods. Besides, we verify the interpretation of our modules through two groups of experiments.

\subsection{Datasets}
\textbf{MSVD} contains 1970 YouTube short video clips. Each video is annotated with multilingual sentences, but we experiment with the roughly 40 captions in English. Similar to the prior work~\cite{Venugopalan2015b}, we separate the dataset into 1,200 train, 100 validation and 670 test videos.

\textbf{MSR-VTT} is another benchmark for video captioning which contains 10,000 open domain videos and each video is annotated with 20 English descriptions. There are 20 simple-defined categories, such as music, sports, movie \etc. we use the standard splits in~\cite{Xu2016} for fair comparison which separates the dataset into 6,513 training, 497 validation and 2,990 test videos.

\textbf{VATEX}\footnote{http://vatex.org/main/index.html} is a most recently released large-scale dataset that reuses a subset of the videos from the Kinetics-600 dataset~\cite{Kay2017} and contains 41,269 videos. Each video is annotated with 10 English and 10 Chinese descriptions. We only utilize English corpora in experiments. Following the official split: 25,991 videos for training, 3,000 videos for validation and 6,000 public test videos for test. Compared with the two datasets mentioned above, the captions are longer and higher-quality, the visual contents are richer and more specific.

\subsection{Implementation Details}

\textbf{Features and Words Preprocessing.} We uniformly sample 28 keyframes/clips for each video and 5 objects for each keyframe. The 1536-D appearance features are extracted by InceptionResNetV2~\cite{Szegedy2017} pretrained on the ImageNet dataset~\cite{Russakovsky2015}. The 2048-D motion features are extracted by C3D~\cite{Hara2018} which is pretrained on the Kinetics-400 dataset, with ResNeXt-101~\cite{Xie2017} backbone. These features are concatenated and projected into hidden space with 512-D. We utilize a ResNeXt-101 backbone based Faster-RCNN pretrained on MSCOCO~\cite{Chen2019a} to extract object features. The object features are captured from the output of FC7 layer without category information and then embedded to 512-D before fed into ORG.

For the sentences longer than 24 words are truncated (30 for VATEX); the punctuation are removed (for VATEX are retained); all words are converted into lower case. We build a vocabulary on words with at least 2 occurrences. We embed the word to 300-D word vector initialized with GloVe by spaCy toolkits.

\textbf{External Language Model Settings.} To guarantee the quality of generated ``soft targets'', we employ the off-the-shelf Bert model provided by \textit{pytorch-transformers} \footnote{https://huggingface.co/transformers/}. Its a bidirectional transformer pretrained using a combination of masked language modeling objective and next sentence prediction on a large corpus comprising the Toronto Book Corpus and Wikipedia. Specifically, the bert-base-uncased model with 12 layers, 768 hidden and 12 self-attention heads is utilized. We then simply fine-tune it on the corpus of corresponding training dataset using Adam~\cite{Kingma2015} optimizer with $3e-5$ learning rate and 128 batch size for 10 epochs. During the captioning model training phase, the parameters of ELM are fixed and we inference the ``soft targets'' of current time step with masking all the words after that.

\textbf{Captioning Model Settings.} The model is optimized by the Adam with a learning rate of 3e-4 and batch size of 128 at training, and we use beam search with size 5 for generation at inference. The two-layers LSTMs used in our decoder have 512 hidden units. The state sizes of both temporal and spatial attentions are set to 512. The dimension of feature vectors in the ORG is 512. We also explore the diverse influences on the system, with the different numbers of top soft targets in TRL and the different number of top collections in P-ORG. In general, top-50 soft targets and top-5 connections are better.

\begin{table*}[]
	\begin{center}
	\resizebox{0.9\textwidth}{!}{%
		\begin{tabular}{@{}c|c|ccc|cccc|cccc@{}}
			\toprule
			\multirow{2}{*}{Models} & \multirow{2}{*}{Year} & \multicolumn{3}{c|}{Features} & \multicolumn{4}{c|}{MSVD} & \multicolumn{4}{c}{MSR-VTT} \\
			&  & Appearence & Motion & Object & B@4 & M & R & C & B@4 & M & R & C \\ \midrule
			SA-LSTM~\cite{Wang2018a} & 2018 & Inception-V4 & - & - & 45.3 & 31.9 & 64.2 & 76.2 & 36.3 & 25.5 & 58.3 & 39.9 \\
			M3~\cite{Wang2018c} & 2018 & VGG & C3D & - & 52.8 & 33.3 & - & - & 38.1 & 26.6 & - & - \\
			RecNet~\cite{Wang2018a} & 2018 & Inception-V4 & - & - & 52.3 & 34.1 & 69.8 & 80.3 & 39.1 & 26.6 & 59.3 & 42.7 \\
			PickNet$^{*}$~\cite{Chen2018} & 2018 & ResNet-152 & - & - & 52.3 & 33.3 & 69.6 & 76.5 & 41.3 & 27.7 & 59.8 & 44.1 \\
			MARN~\cite{Pei2019} & 2019 & ResNet-101 & C3D & - & 48.6 & 35.1 & 71.9 & 92.2 & 40.4 & 28.1 & 60.7 & 47.1 \\
			SibNet~\cite{Liu2018} & 2019 & GoogleNet & - & - & 54.2 & 34.8 & 71.7 & 88.2 & 40.9 & 27.5 & 60.2 & 47.5 \\
			OA-BTG~\cite{Zhang2019} & 2019 & ResNet-200 & - & Mask-RCNN & \textbf{56.9} & 36.2 & - & 90.6 & 41.4 & 28.2 & - & 46.9 \\
			GRU-EVE~\cite{Aafaq2019} & 2019 & InceptionResnetV2 & C3D & YOLO & 47.9 & 35.0 & 71.5 & 78.1 & 38.3 & 28.4 & 60.7 & 48.1 \\
			MGSA~\cite{Chen2019b} & 2019 & InceptionResnetV2 & C3D & - & 53.4 & 35.0 & - & 86.7 & 42.4 & 27.6 & - & 47.5 \\
			POS+CG~\cite{Wang_2019_ICCV} & 2019 & InceptionResnetV2 & OpticalFlow & - & 52.5 & 34.1 & 71.3 & 88.7 & 42.0 & 28.2 & 61.6 & 48.7 \\
			POS+VCT~\cite{Hou_2019_ICCV} & 2019 & InceptionResnetV2 & C3D & - & 52.8 & 36.1 & 71.8 & 87.8 & 42.3 & \textbf{29.7} & \textbf{62.8} & 49.1 \\ \midrule
			ORG-TRL & Ours & InceptionResnetV2 & C3D & FasterRCNN & 54.3 & \textbf{36.4} & \textbf{73.9} & \textbf{95.2} & \textbf{43.6} & 28.8 & 62.1 & \textbf{50.9} \\ \bottomrule
		\end{tabular}%
	}
	\end{center}	
	\caption{Performance comparisons on MSVD and MSR-VTT benchmarks. The best results and corresponding features are listed.}
	\label{tab:state-of-the-art}
	\vspace{-0.5cm}
\end{table*}
\subsection{Performance Comparison}

To evaluate the effectiveness of our models, we compare our model with state-of-the-art models listed in Tab.\ref{tab:state-of-the-art}. Due to diverse modalities for video captioning, we list the models that only contain visual modalities \ie appearance, motion and object features. Even so, it's also hard to achieve a completely fair comparison because of different feature extraction methods. Therefore, we try to employ the same feature extractors and preprocessing as the most recent models.

The quantitative results in Tab.\ref{tab:state-of-the-art} illustrate our model gets significant improvement on MSVD and MSR-VTT datasets, which verifies the effectiveness of our proposed methods. Specifically, compared with GRU-EVE, MGSA, POS+CG and POS+VCT using the same features as ours, which demonstrate the superior performance without the effects of features. The remarkable improvement under CIDEr on both datasets demonstrates the ability to generate novel words of our model. Since the mechanism of CIDEr is to punish the often-seen but uninformative n-grams in the dataset. This phenomenon verifies that our model captures the detailed information from videos and acquires wealthy knowledge via ELM.

Moreover, we compare our model with the existing video captioning models that use detailed object information. GRU-EVE tries to derive high-level semantics from an object detector to enrich the representation with spatial dynamics of the detected objects. OA-BTG applies a bidirectional temporal graph to capture temporal trajectories for each object. However, these two methods ignore the relationship between objects. Our ORG method achieves better performances than OA-BTG on MSR-VTT in Tab.\ref{tab:state-of-the-art}, which illustrates the benefits of object relations. Note that, POS+VCT achieves higher scores under METEOR and ROUGE-L on MSR-VTT, and these are probably caused by the reason that their POS method can learn the syntactic structure representation.

Besides, we also report the results of our model on the public test set of recent published VATEX dataset as shown in Tab.\ref{tab:vatex_english}. These results come from the online test system. Compared with the baseline model, we train the model on English corpus without sharing Encoder and Decoder.

\begin{table}[]
	\begin{center}
	\resizebox{0.4\textwidth}{!}{%
		\begin{tabular}{@{}l|cccc@{}}
			\toprule
			Model & B@4 & M & R & C \\ \midrule
			Shared Enc~\cite{Wang2019} & 28.9 & 21.9 & 47.4 & 46.8 \\
			Shared Enc-Dec~\cite{Wang2019} & 28.7 & 21.9 & 47.2 & 45.6 \\ \midrule
			Baseline(Ours) & 30.2 & 21.3 & 47.9 & 44.6 \\
			Baseline+ORG(Ours) & 31.5 & 21.9 & 48.7 & 48.8 \\
			Baseline+TRL(Ours) & 31.5 & 22.1 & 48.7 & 49.3 \\
			Baseline+ORG+TRL(Ours) & \textbf{32.1} & \textbf{22.2} & \textbf{48.9} & \textbf{49.7} \\ \bottomrule
		\end{tabular}%
	}
	\end{center}
	\caption{The results of the VATEX online evaluation system.}
	\label{tab:vatex_english}
	\vspace{-0.3cm}
	
\end{table}

\begin{table}[]
	\begin{center}
	\resizebox{0.45\textwidth}{!}{
		\begin{tabular}{@{}cc|cccc|cccc@{}}
			\toprule
			\multicolumn{2}{c|}{Methods} & \multicolumn{4}{c|}{MSVD} & \multicolumn{4}{c}{MSR-VTT} \\
			ORG & TRL & B@4 & M & R & C & B@4 & M & R & C \\ \midrule
			$\times$ & $\times$ & 53.3 & 35.2 & 72.4 & 91.7 & 41.9 & 27.5 & 61.0 & 47.9 \\
			\checkmark & $\times$ & 54.0 & 36.0 & 73.2 & 94.1 & 43.3 & 28.4 & 61.5 & 50.1 \\
			$\times$ & \checkmark & 54.0 & 36.0 & 73.7 & 93.3 & 43.2 & 28.6 & 61.7 & 50.4 \\
			\checkmark & \checkmark & \textbf{54.3} & \textbf{36.4} & \textbf{73.9} & \textbf{95.2} & \textbf{43.6} & \textbf{28.8} & \textbf{62.1} & \textbf{50.9} \\ \bottomrule
		\end{tabular}%
	}
	\end{center}
	\caption{Ablation Studies of the ORG and the TRL on MSVD and MSR-VTT benchmarks.}
	\label{tab:ablation satudies}
	\vspace{-0.5cm}
\end{table}

\subsection{Ablation Experiments}

\textbf{Effectiveness of each component}. We design 4 control experiments to demonstrate the effectiveness of the proposed ORG module and TRL. Tab.\ref{tab:ablation satudies} gives the control results on the testing set of MSVD and MSR-VTT datasets. The baseline model only applies appearance and motion features, and the same encoder-decoder architecture as mentioned above except without object encoder. It follows the Cross-Entropy criterion, and the results are shown in the first row of the table. Compared with the baseline model, both ORG and TRL achieve improvement when added alone. The combination of two methods can further enhance the performance which is illustrated as the last row. 

\begin{table}[]
	\begin{center}
	\resizebox{0.37\textwidth}{!}{%
		\begin{tabular}{@{}cl|c|cccc@{}}
			\toprule
			\multicolumn{2}{c}{Methods} & Top-k & B@4 & M & R & C \\ \midrule
			\multicolumn{2}{c}{Baseline(B)} & - & 41.9 & 27.5 & 61.0 & 47.9 \\
			\multicolumn{2}{c}{B+P-ORG} & - & 43.1 & 28.3 & 61.4 & \textbf{50.4} \\
			\multicolumn{2}{c}{B+C-ORG} & 5 & \textbf{43.3} & \textbf{28.4} & \textbf{61.5} & 50.1 \\ \midrule
			\multicolumn{2}{c}{B+C-ORG} & 1 & 42.4 & 28.4 & 61.2 & 49.3 \\
			\multicolumn{2}{c}{B+C-ORG} & 20 & 42.9 & 28.4 & 61.8 & 50.0 \\
			\multicolumn{2}{c}{B+C-ORG} & All & 42.8 & 28.2 & 61.2 & 49.3 \\ \bottomrule
		\end{tabular}%
	}
	\end{center}	
	\caption{Ablation for two kinds of ORGs (top-half), and performance comparisons of the C-ORG with different top-k objects (bottom-half) on MSR-VTT.}
	\label{tab:org-ablation}
	\vspace{-0.5cm}
\end{table}

\begin{figure*}
	\centering
	\includegraphics[width=1\linewidth]{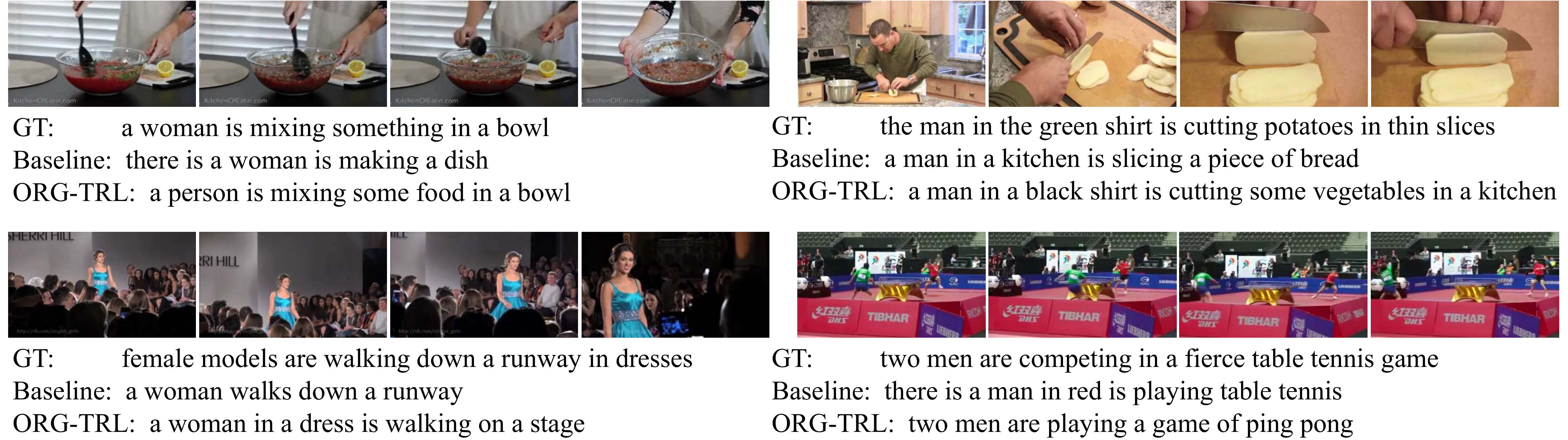}
	\caption{Examples of generations on MSR-VTT with the baseline model and our proposed ORG-TRL system.}
	\label{fig:overall effect}
	\vspace{-0.5cm}
\end{figure*}

\textbf{The evaluation of ORG.} We explore two proposed ORGs: the P-ORG and the C-ORG, and we connect top-5 object nodes for C-ORG. The top half of Tab.\ref{tab:org-ablation} demonstrates that C-ORG is better than P-ORG. It is probable that P-ORG can get more comprehensive information from the whole video than P-ORG. Moreover, both ORGs achieve significant improvement compared with the baseline model, which attributes to the association between objects.
We also explore the effect of different Top-k for the C-ORG which is listed in the bottom half of Tab.\ref{tab:org-ablation}. ``All'' means each node acquires information from all nodes. We find that the highest performances are achieved at the sweat point when $k=5$. A proper explanation is that, when $k$ is too small, there are not enough related objects to update the relation of node; when k is too large, a few unrelated nodes will be introduced and bring noise.

 \begin{figure}
	\centering
	\includegraphics[width=0.9\linewidth]{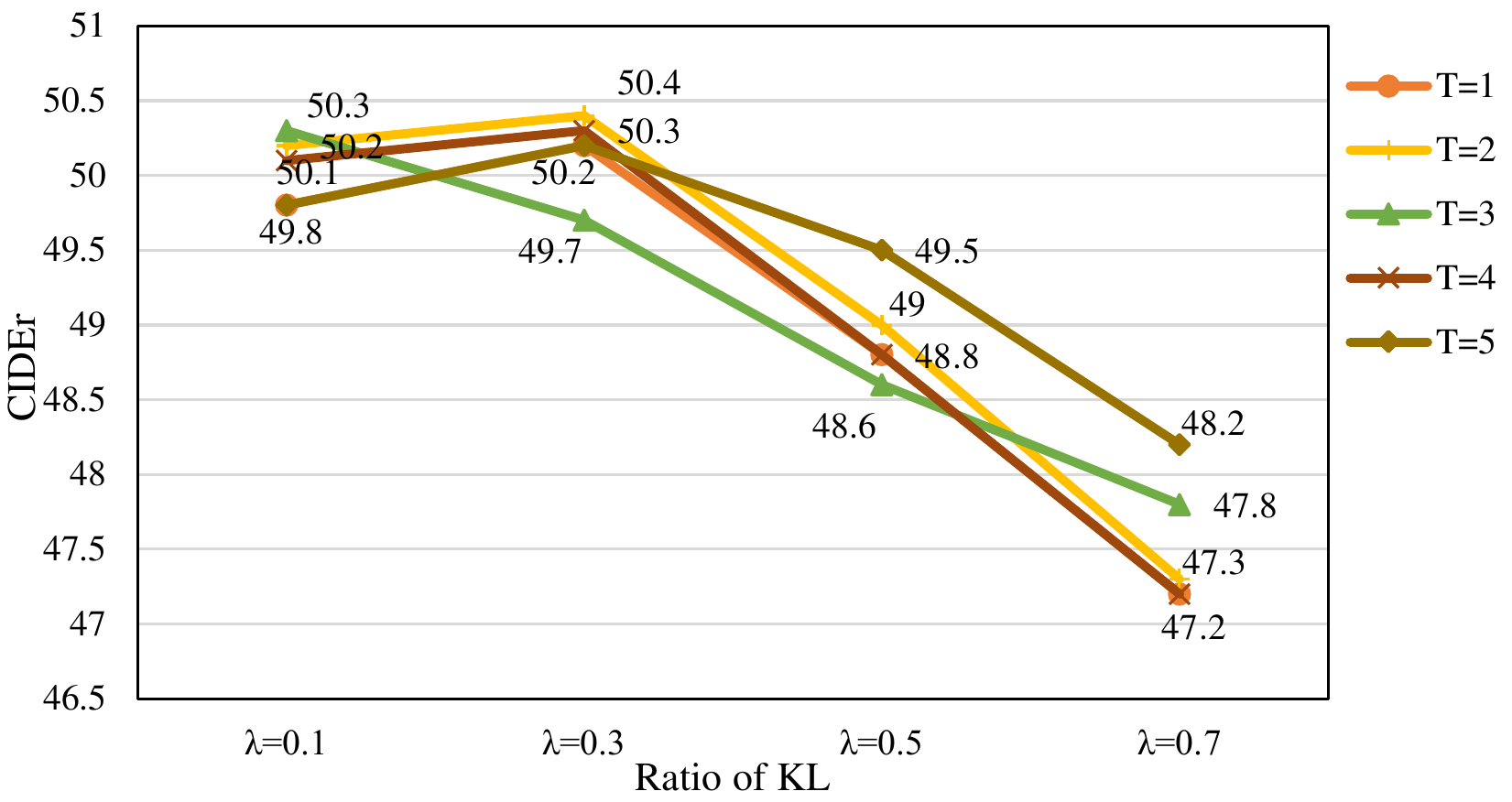}
	\caption{Analysis of different temperatures of ELM and different ratios of KL-loss on MSR-VTT.}
	\label{fig:temp and kl ratio}
	\vspace{-0.5cm}
\end{figure}

\begin{figure}
	\centering
	\includegraphics[width=\linewidth]{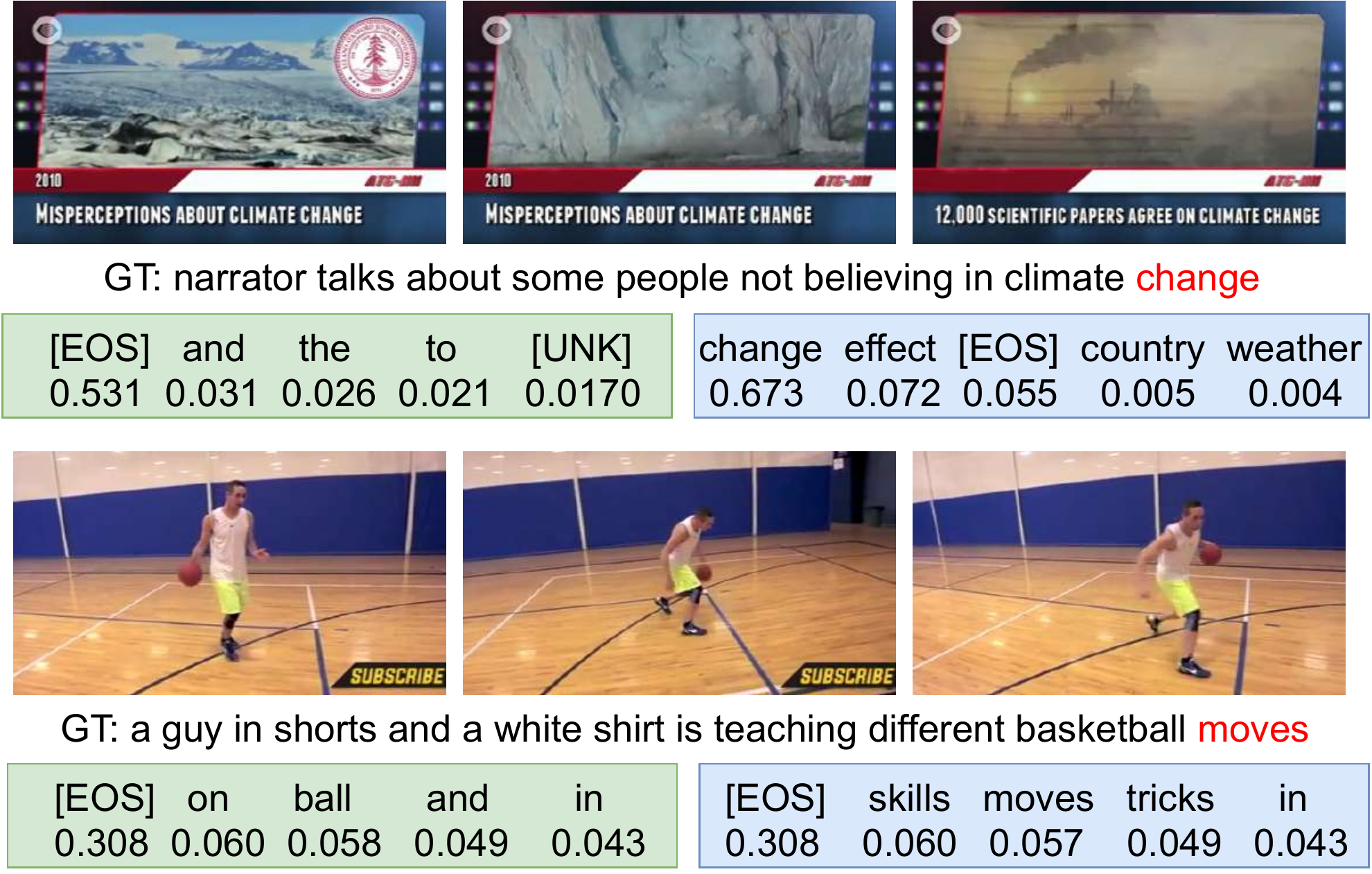}
	\caption{Two instances of the baseline model and baseline+TRL model in inference. The red word is the word to be predicted. The left-green box is the prediction of baseline model; the right-blue box is under the guidance of TRL.}
	\label{fig:trl_analysis}
	\vspace{-0.7cm}
\end{figure}

\textbf{The evaluation of TRL.} We analyze the effect of different ELM temperatures $T_e$ and different ratios of KL-loss $\lambda$. Fig.\ref{fig:temp and kl ratio} illustrates the performances on CIDEr: If $T_e$ is too low, the distribution of soft targets is sharp, thus large noise will be introduced if top-one is not the content related word. Otherwise, the distribution is too smooth to reflect the importance of soft targets. On the other hand, the weight of $\lambda$ reflects the degree of the TRL: the generation will deviate from the content of the video itself if $\lambda$ is too high; it plays no role if too low. Fig.\ref{fig:trl_analysis} shows comparisons of the intermediate states of the baseline model and our TRL based model at inference time, and two models are trained with the same epoch: the red word is the next word to be predicted; the green box and blue box show the predictions and their probabilities of baseline model and our TRL method respectively. See the first clip, ``climate change'' is a very common noun phrase, but it rarely appears in the caption task. As shown in the second clip, the caption model can predict various proper combinations after ``basketball'' according to the sentence context. Moreover, the most of words are relative with video content. Our TRL method can help the model to learn some common matches and content-related words. To some extent, it effectively alleviates the long-tailed problem of video captioning task. We also experiment various top-k soft targets, see the appendix for detail.

\subsection{Qualitative Analysis}
We show some examples in Fig.\ref{fig:overall effect}. It can be seen, the content of captions generated by our model is richer than the baseline model, and more activity associations are involved. For instance, the example at top-left shows that the baseline model can only understand the general meaning of the video. By contrast, our model can recognize more detailed objects, and the relation ``mixing'' between ``person'' and ``food'', even the position ``in a bowl''. The rest of the examples have similar characteristics.
\section{Conclusion}
In this paper, we have proposed a complete system, which contains a novel model and a training strategy for video captioning. By constructing relational graph between objects and performing relational reasoning, we can acquire more detailed and interactive object features. Furthermore, the novel TRL introduces external language model to guide the caption model to learn abundant linguistic knowledge, which is the supplement of the common TEL. Our system has achieved competitive performances on MSVD, MSR-VTT and VATEX datasets. The experiments and visualizations have demonstrates the effectiveness of our methods.\\
\textbf{Acknowledgements} This work is supported by NSFC-general technology collaborative Fund for basic research (Grant No.U1636218, U1936204), Natural Science Foundation of China (Grant No.61751212, 61721004, U1803119), Beijing Natural Science Foundation (Grant No.L172051, JQ18018), CAS Key Research Program of Frontier Sciences (Grant No.QYZDJ-SSW-JSC040), CAS External cooperation key project, and National Natural Science Foundation of Guangdong (No.2018B030311046). Bing Li is also supported by Youth Innovation Promotion Association, CAS.

{\small
	\bibliographystyle{ieee_fullname}
	\bibliography{video_captioning}
}
\end{document}